\documentclass[manuscript,screen,nonacm]{acmart}
\AtBeginDocument{%
  \providecommand\BibTeX{{%
    \normalfont B\kern-0.5em{\scshape i\kern-0.25em b}\kern-0.8em\TeX}}}

\setcopyright{acmlicensed}
\copyrightyear{2018}
\acmYear{2018}
\acmDOI{XXXXXXX.XXXXXXX}

\acmConference[Conference acronym 'XX]{Make sure to enter the correct
  conference title from your rights confirmation emai}{June 03--05,
  2018}{Woodstock, NY}
\acmBooktitle{Woodstock '18: ACM Symposium on Neural Gaze Detection,
 June 03--05, 2018, Woodstock, NY} 
\acmISBN{978-1-4503-XXXX-X/18/06}

\usepackage{tabularx}
\usepackage{color, soul}
\usepackage{pifont}
\usepackage{caption}
\usepackage{subcaption}
\usepackage{xcolor}
\newlength\MAX  

\newcolumntype{b}{X}
\newcolumntype{s}{>{\hsize=.5\hsize}X}
\newcolumntype{t}{>{\hsize=.25\hsize}X}
\newcolumntype{m}{>{\hsize=.10\hsize}X}

\begin{document}

\title{VITA: A Multi-modal LLM-based System for Longitudinal, Autonomous, and Adaptive Robotic Mental Well-being Coaching}

\renewcommand{\shorttitle}{VITA system}


\author{Micol Spitale}
\authornote{This work was undertaken and finalised while M. Spitale was a postdoctoral researcher at the University of Cambridge.}
\affiliation{%
  \institution{Politecnico di Milano}
  \city{Milan}
  \country{Italy}
}
\email{micol.spitale@polimi.it}
\orcid{0000-0002-3418-1933}

\author{Minja Axelsson}
\affiliation{%
  \institution{University of Cambridge}
  \city{Cambridge}
  \country{UK}}
\email{mwa29@cam.ac.uk}

\author{Hatice Gunes}
\affiliation{%
  \institution{University of Cambridge}
  \city{Cambridge}
  \country{UK}}
\email{hatice.gunes@cl.cam.ac.uk}

\renewcommand{\shortauthors}{Spitale, et al.}

\begin{abstract}
 Recently, several works have explored if and how robotic coaches can promote and maintain mental well-being in different settings. 
 However, findings from these studies revealed that these robotic coaches are not ready to be used and deployed in real-world settings due to several limitations that span from technological challenges to coaching success. To overcome these challenges, this paper presents VITA, a novel multi-modal LLM-based system that allows robotic coaches to autonomously adapt to the coachee's multi-modal behaviours (facial valence and speech duration) and deliver coaching exercises in order to promote mental well-being in adults. We identified five objectives that correspond  to the challenges in the recent literature, and we show how the VITA system addresses these via experimental validations that include one in-lab pilot study ($N$ = 4) that enabled us to test different robotic coach configurations (pre-scripted, generic, and adaptive models) and inform its design for using it in the real world, and one real-world study ($N$ = 17) conducted in a workplace over 4 weeks. 
 Our results show that: (i) coachees perceived the VITA adaptive and generic configurations more positively than the pre-scripted one, and they felt understood and heard by the adaptive robotic coach, (ii) the VITA adaptive robotic coach kept learning successfully by personalising to each coachee over time and did not detect any interaction ruptures during the coaching, (iii) coachees had significant mental well-being improvements via the VITA-based robotic coach practice. 
 The code for the VITA system is openly available via \url{https://github.com/Cambridge-AFAR/VITA-system}.
 
\end{abstract}





\maketitle

\begin{table*}[htb!]
    \footnotesize
    \centering
    \begin{tabularx}{\textwidth}{X|X|X}
    \toprule
         \textbf{Limitations of past works (L)}& \textbf{Objectives of this work (O)}& \textbf{Experimental validation (EV)}\\
         \midrule
         \textbf{(L1)} Coachees reported negative perception and feedback about the robotic coach because they found it \textit{not} interactive nor responsive \cite{spitale_axelsson_kara_gunes_2023} or "glitchy" \cite{jeong2023deploying}. & \textbf{(O1)} Making the robotic coach more interactive and responsive to improve coachees' perception towards the robotic coach & 
          NARS questionnaire and GEQ results of the pilot study (see Section \ref{sec:pilot}, \textbf{EV1.1}) and the real-world study (see Section \ref{sec:quest-results}, \textbf{EV1.2}), and  coachees's interview comments on the perception of the robotic coach (see Section \ref{sec:int-results}, \textbf{EV1.3})\\
         \midrule

         \textbf{(L2)} The robotic coach practice was \textit{not personalised} longitudinally, in contrast to recommendations of \cite{axelsson2022robots} and \cite{jeong2023deploying} and coachees expected it to "adapt and personalise more" \cite{spitale2023robotic}. & \textbf{(O2)} Implementing and evaluating an adaptive robotic coach to better match with the coachee's behaviour over time & 
          Reward increasing over time in the pilot study (see Figure \ref{fig:pilot_reward}, \textbf{EV2.1}) and in the real-world study (see Figure \ref{fig:realworld} and Section \ref{sec:vita-results}, \textbf{EV2.2}), coachees' perception of adaptation (see Section \ref{sec:real-world}), their overall impression (see Table \ref{tab:overall-impression}, \textbf{EV2.3}) and   coachees's interview comments on the perception of the robotic coach (see Section \ref{sec:int-results}, \textbf{EV2.4})\\
         
         \midrule

         \textbf{(L3)} Coachees experienced \textit{several interaction ruptures} during robotic coaching \cite{spitale_axelsson_kara_gunes_2023} that can negatively affect the coach-coachee alliance. & \textbf{(O3)} Recognising the occurrences of interaction ruptures via automatic detection during robotic coaching to guarantee the success of the practice & 
         Interaction ruptures detected in the real-world study (see Section \ref{sec:quest-results}, \textbf{EV3.1}), and coachees's interview comments on the flow of the coaching practice (see Section \ref{sec:int-results}, \textbf{EV3.2}) \\
         \midrule
         
         \textbf{(L4)} Coachees found the conversational capabilities of the robotic coach were limited, as in \cite{spitale2023robotic, jeong2020robotic}, and they did not feel understood or listened to.  & \textbf{(O4)} Embedding the robot with the advance large language model (LLM) module to imitate human coach behaviours&
         Coachees's interview comments on conversational capabilities of the robotic coach (see Section \ref{sec:int-results}, \textbf{EV4.1}) \\
         
         \midrule
         
         \textbf{(L5)} Coachees' mental well-being \textit{did not significantly} improve via robotic coaching in the long term \cite{spitale2023robotic}. & \textbf{(O5)} \textit{Significant} improvement in the coachees' mental well-being to demonstrate the success of the robotic coaching & 
         \textit{Significant} RPWS scale results in the real-world study (see Figure \ref{fig:realworld} and Section \ref{sec:quest-results}, \textbf{EV5.1}), and coachees's interview comments on the benefits of the coaching practice (see Section \ref{sec:int-results}, \textbf{EV5.2}) \\
         \bottomrule
         
    \end{tabularx}
    
    \caption{Limitations of past works (L), Objectives to address those limitations (O), 
    and Experimental Validation of the VITA system designed to achieve those objectives (EV).}
    \label{tab:motivation}
\end{table*}

\section{Introduction}




Mental health and well-being promotion and maintenance have been paramount goals for the World Health Organization (WHO)\footnote{https://www.who.int/health-topics/mental-health} in the last decade, given the increase of mental health conditions worldwide. Despite the rise of mental healthcare needs, the treatment coverage remains extremely low leading to a call for innovative mental healthcare solutions.
To address the limited access to healthcare systems, several works \cite{spitale2022affective, matheus2022social, jeong2023deploying} have recently investigated if and how robotic coaches can be used to promote mental well-being in different contexts, e.g., workplace \cite{spitale2023robotic}, student accommodations \cite{jeong2020robotic} etc. 
These studies show that the robotic technologies have the potential of providing a dedicated space and time for a coaching session that smartphone-based apps cannot 
\cite{axelsson2023robotic}.
However, findings from these studies revealed that robotic coaches are still very far from being used and deployed in the real world as fully autonomous coaching agents for long term use due to several limitations that span from the technological challenges to the success of the practice.

Overall, findings in \cite{jeong2020robotic, spitale2023robotic} showed that coachees expect from the robotic coaches more than they can currently deliver. The coachees complained about the lack of interactivity and responsiveness of the robotic coach \cite{axelsson2023robotic, spitale2023robotic}, or the lags in robot response and behaviour, and various glitches \cite{jeong2023deploying} (\textbf{Limitation 1}, \textbf{L1}). 
Findings from past works also showed that coachees expected the robotic coach to "adapt and personalise more" \cite{spitale2023robotic}. However, none of the past works have explored the use of personalised and adaptive robotic coaching deploying it in the long term despite the recommendations distilled by \cite{axelsson2022robots} and \cite{jeong2023deploying} (\textbf{L2}).
Additionally, coachees experienced several interaction ruptures (i.e., coachees felt awkward and/or the robot was making a mistake, e.g., interrupting the coachees \cite{spitale_axelsson_kara_gunes_2023}) that may have also jeopardized the coach-coachee alliance, another important factor for the success of the coaching practice \cite{de2017coaching} (\textbf{L3}).
Coachees also highlighted the importance of the conversational capabilities of the robotic coaches. They reported that the dialogue with the robot was very limited \cite{jeong2023deploying} and that they did not feel understood or heard \cite{spitale2023robotic} (\textbf{L4}). 
Past works have demonstrated the efficacy of robotic coaching (i.e., leading to an improvement of coachees' mental well-being) when delivering coaching exercises in short-term interactions (e.g., after 7-days interaction \cite{jeong2020robotic}). However, the very few papers  
 \cite{bodala2021teleoperated, spitale2023robotic} that explored their efficacy in the long term did not find any significant improvements in the coachees' mental well-being (\textbf{L5}). The factors that could have negatively influenced the success of the practice may be multiple, including the technical limitations of the robots highlighted above.

In this paper, we present \textbf{VITA} (“life” in latin), a novel multi-modal LLM-based system for adaptive mental well-being coaching that allows a robotic coach to autonomously adapt to each coachee and deliver coaching exercises to promote mental well-being in adults by addressing the above-mentioned limitations as reported in Table \ref{tab:motivation}.
The main objectives of this work are the following. 
 First, we design and develop the robotic coach to be interactive (i.e., multi-modal two-way interaction between the coach and the coachee) and responsive (i.e., the robotic coach can respond to the coachee by analysing both the coachees' verbal and non-verbal behaviours) and we evaluate the coachees' perceptions towards the robotic coach (\textbf{Objective 1}, \textbf{O1} addressing L1). 
 Second, we develop and implement an adaptive and personalised model to manage the dialogue flow of the coaching practice and evaluate its advantages by comparing it with a non-adaptive model  (\textbf{O2} addressing L2). 
 Then, we implement an automatic model based on machine learning to recognise interaction ruptures during the interaction with the robotic coach, and we embed it into the robotic coaching system to evaluate whether ruptures occur during the coaching practice (\textbf{O3} addressing L3). 
We then leverage the advancements in large language models (LLMs) and embed a ChatGPT-based conversational module using prompt engineering into the robotic coach system to deliver positive psychology exercises (\textbf{O4} addressing L4).
 Finally, we evaluate if coachees who interacted in the long-term with a fully autonomous and adaptive robotic coach reported significant improvements in their mental well-being (\textbf{Objective 5}, \textbf{O5} addressing L5).

To simultaneously address all these objectives in a single work, we conducted various evaluations by undertaking a pilot in-lab study -- that involved 4 participants -- to evaluate the differences between \textit{pre-scripted} (i.e., the robotic coach follows a pre-defined sequence of utterances to speak aloud to the coachee no matter what the coachee says), \textit{generic} (i.e., the robotic coach understands via natural language processing what the coachee says and responds accordingly without personalising the selection of the next line of the dialogue to coachee's behaviour), and \textit{adaptive} (i.e., the robotic coach understands via natural language processing what the coachee says and personalises to the coachee's behaviour by selecting the next line of the dialogue flow accordingly) models. This initial study informed us that the adaptive model led to a more personalised interaction than the pre-scripted and generic models. Based on this finding we embedded the robotic coach system with the adaptive model to manage the dialogue flow during the coaching practice delivered by the robot. We deployed our implementation in a long-term real-world study -- that involved 17 participants at a tech company -- to evaluate the capabilities of the VITA-based robotic well-being coaching. 

The main contributions of this paper are two-fold. First, we provide the community with a novel and open-source system named VITA to design and develop an autonomous and adaptive robotic coach that can deliver mental well-being practices by adapting to each coachee. Our second contribution is the evaluation of the VITA-based robotic coach via pilot and long-term real-world studies to successfully deliver mental well-being coaching practices. Specifically, we found that (i) coachees perceived the adaptive and generic configurations more positively than the pre-scripted one, and they felt understood and heard by the robotic coach, (ii) the adaptive robotic coach kept learning successfully by personalising to each coachee over time and did not detect any interaction ruptures during the coaching practice, and (iii) coachees reported significant mental well-being improvements via the robotic coach practice. 

\section{Related Work}

\subsection{Robotic coaches for mental well-being}
 Just a handful of papers have researched the use of robotic coaches to foster mental well-being, e.g., \cite{axelsson2023robotic, spitale2022affective, spitale2023robotic, jeong2023deploying, bodala2021teleoperated, matheus2022social, shi2023evaluating, abbasi2022can}. 
 \citet{jeong2020robotic} conducted a longitudinal study which utilized Jibo robots to provide positive psychology interventions to students in home settings over a period of seven days, and reported that participants experienced a better sense of well-being, improved mood and a readiness to change, while gradually developing an affinity for the robot. 
\citet{shi2023evaluating} investigated the effect of physical embodiment and personalisation on the user-perceived quality of text-to-speech (TTS) voices for mindfulness. Their results showed that the user-personalized TTS voices were able to perform nearly as well as human voices, indicating that user personalisation could be a powerful approach to raise user perception of TTS voice quality.
\citet{spitale2023robotic, spitale_axelsson_kara_gunes_2023} conducted a study involving employees of a tech company to interact with two different forms of robotic coaches that delivered positive psychology exercises over 4 weeks. Their results showed that the robot form may impact the perception of the coachees towards the robotic coach. \citet{jeong2023robotic} explored the companion-like robotic coach behaviour during well-being therapy by comparing the assistant, coach, and companion roles of the robot. Their results showed that the companion robot was the most effective in building a positive therapeutic alliance with its users. 

\subsection{HRI tools and systems}

There are notable efforts within the HRI community to create tools for composing social human-robot interactions \cite{rajendranframework,7451766, siepmann2013behavior}, including robot specific tools, (e.g., Choreograph \cite{pot2009choregraphe}), content authoring tools (e.g., \cite{coronado2020visual}), and dialogue management tools (e.g., \cite{huang2012robot}). However, these tools are not multimodal, open-source, robot-agnostic, composable, and modular  \--- crucial factors for adoption and wide usage by the HRI community \cite{spitale2021composing}. One representative example that addresses the aforementioned open issues is HARMONI \cite{spitale2021composing} \--- an open-source, robotic-agnostic, modular and multi-modal tool to help researchers compose social human-robot interactions. 
Such tools are extremely beneficial when maintained and supported by the HRI community as they enable researchers to quickly implement a new interaction or deploy a new robot for a research study without wasting time to reinvent the wheel. 

However, such tools are usually implemented for generic use. On the contrary, robotic \textit{systems} (e.g., \cite{odabasi2022refilling}) refer to the development and, if applied, the use of the above-mentioned tools, to specific application scenarios, such that they "synthesise underlying techniques to achieve system-level HRI behavior" \footnote{https://humanrobotinteraction.org/2024/fullpaper/}. 
To this aim, \citet{taylor2022regroup} presented the Robot-Centric Group Detection and Tracking System (REGROUP) as a new system that enables robots to detect and track groups of using a crowd-aware, tracking by-detection approach that has been made publicly available. Analogously, \citet{nanavati2022not} introduced a novel system that relies on informed direction selection to avoid obstacles and traverse an hallway, and periodic human help to charge, evaluating it in the wild and shared their code open source. 
However, none of the existing robotic systems have focussed on robotic well-being coaching.
This paper presents a novel adaptive robotic system named VITA that leverages the HARMONI tool to promote code re-use and advances the current HRI state of the art by enabling the autonomous delivery of well-being practices.

\subsection{Adaptation and personalization in HRI}
Within the HRI literature, very recently a number of works, e.g., \cite{axelsson2023you, gillet2022learning, churamani2022continual, mcquillin2022learning}, have started to investigate and evaluate the use of adaptive and personalised robots by showing that adaptive configurations are very promising for robotic applications \cite{rossi2017user}. For example, \citet{axelsson2023you} presented a fully automated system for building adaptive presentations for embodied agents. They evaluated the system involving 43 participants who interacted with the adaptive system, and their results demonstrated that the user preferred the adaptive system. Analogously, \citet{gillet2022learning} explored how learning robot gaze behaviors (via reinforcement learning) can balance human participation in conversational interactions. 
Their results showed that the proposed reward for the reinforcement learning approach enabled the robot to encourage participants to take more turns.
Only a few works have explored the applications of adaptation and personalisation in robots in the context of mental well-being. \citet{churamani2022continual} proposed a novel system for well-being coaching by utilising continual learning to personalise the robotic coaching to each user, and compared static, adaptive, and personalised versions of the robotic coach. 
Their results showed that overall the users tend to prefer the robotic coach with continual personalisation.

However, none of the previous works have created \textbf{long-term autonomous adaptation} and/or \textbf{personalisation} strategies for \textbf{robotic mental well-being coaching}. 

\section{VITA System}
\label{sec:system}

This section introduces the VITA system providing a general overview and a detailed description of the main components embedded in the robotic coach 
that have been used and tested during the pilot and real-world studies to address the objectives listed in Table \ref{tab:motivation}. 

\subsection{Overview}

\begin{figure}
    \centering
    \includegraphics[width = 0.95\columnwidth]{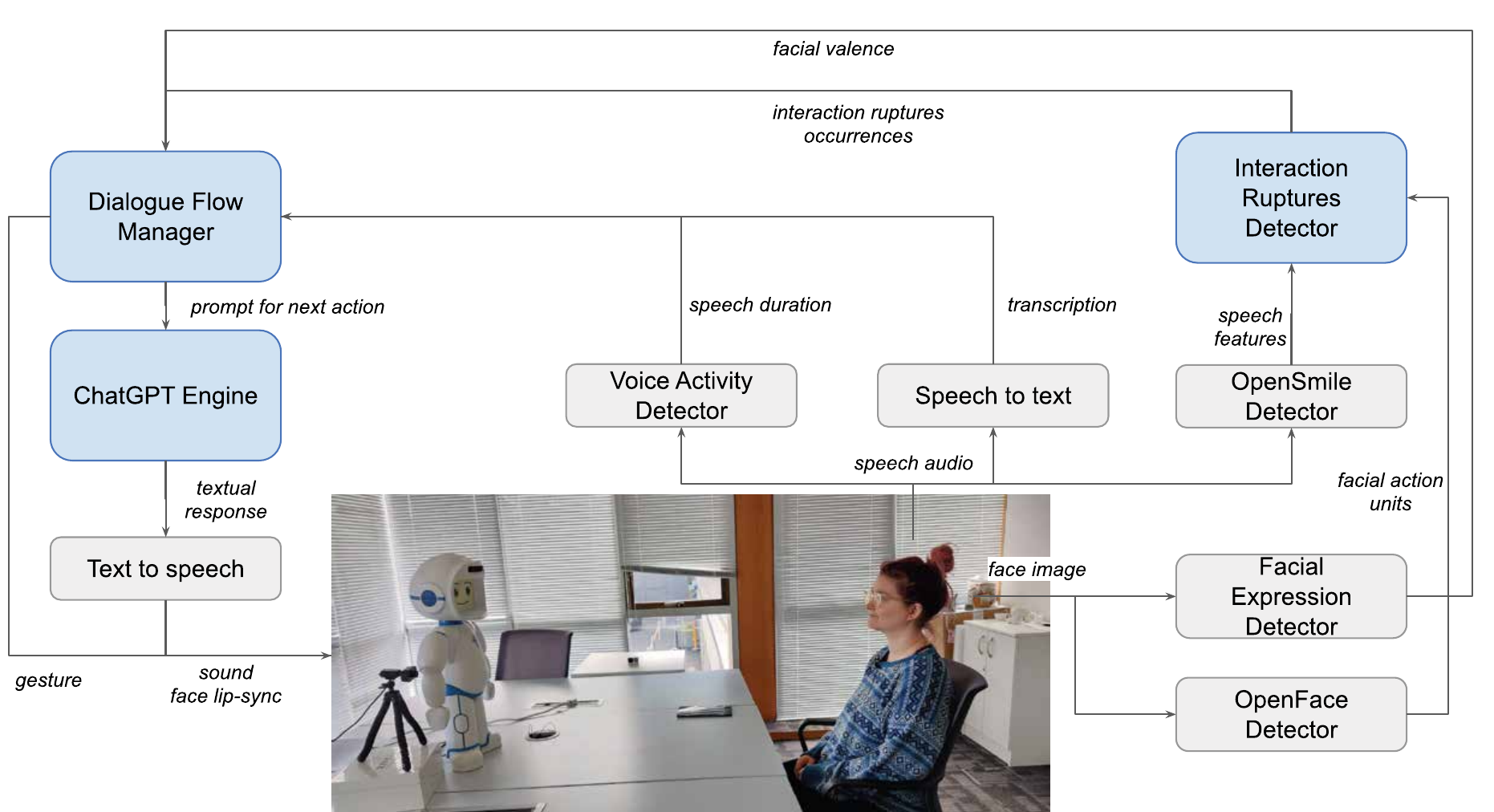}
    \caption{The components of the VITA system for adaptive robotic mental well-being coaching.}
    \label{fig:vita}
\end{figure}

VITA is based on a \textit{behaviour tree} structure which runs at a frequency of 10Hz and is managing all the VITA components from a high level. The behaviour tree, inspired from \cite{spitale2021composing} -- in which the authors presented a very simple pre-scripted dialogue-based flow -- coordinates reading the data from the sensors (microphone and camera of the robot), running the detectors continuously (e.g., facial expression, interaction ruptures etc.), calling external services (e.g., OpenAI API for natural language processing), and activating the actuators (e.g., speaker and motors of the robot) when a decision about the dialogue flow is made. 

A walk-through of the interaction is depicted in Figure \ref{fig:vita} and described as follows. To begin, the sensors (microphone and camera) and the detectors (openSMILE, OpenFace, facial expression, voice activity, and \textit{interaction ruptures}) are activated. The \textit{dialogue flow manager} initiates the interaction, and the robot introduces the positive psychology exercise and asks the first question. Then, the coachee is required to answer the questions posed by the robotic coach and the detectors analyse the raw audio and video data published by the microphone and the camera of the robot. After that, the detectors send their outputs to the dialogue flow manager which decides, on the basis of the multi-modal behavioural data collected from the detectors, which dialogue action to take next in the coaching practice and sends the corresponding prompt to the \textit{ChatGPT engine}. This produces its textual response and sends it to the actuators that synthesise the text and generate the robotic coach behaviour (i.e., sound, facial lip-sync and gesture).

\begin{figure}[htb!]
    \centering
    \includegraphics[width=\columnwidth]{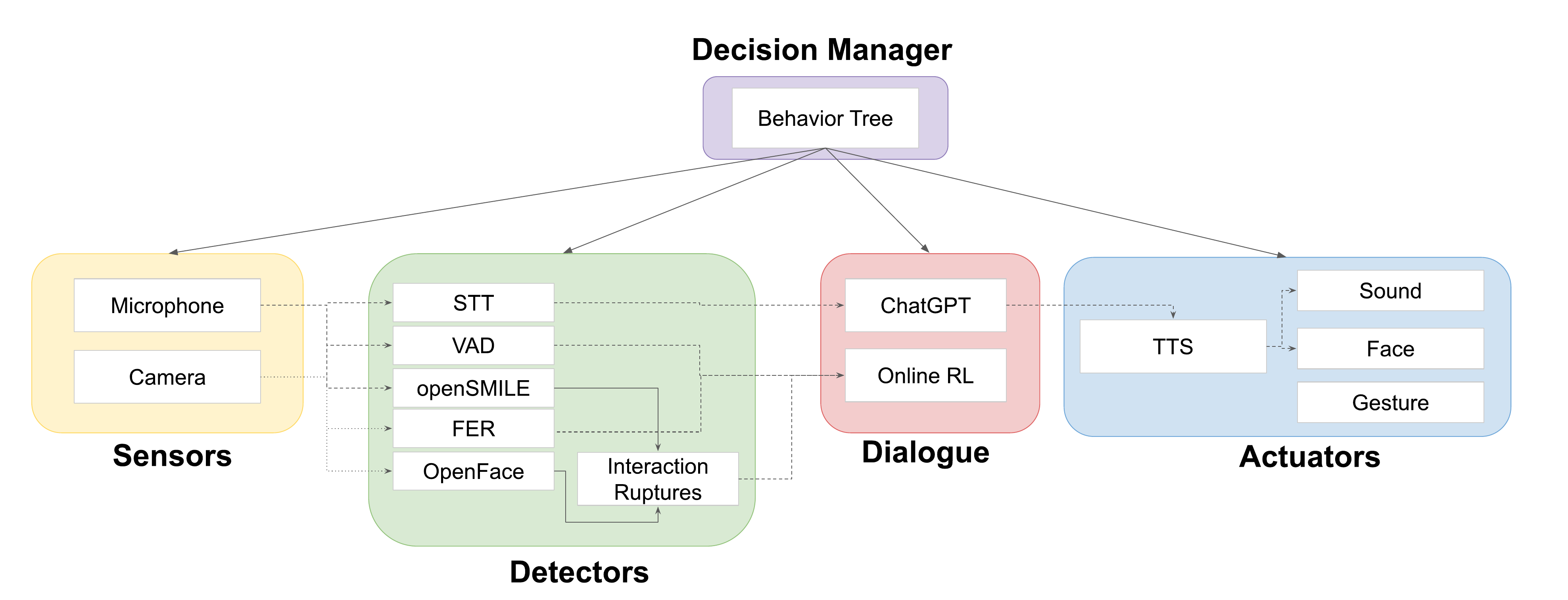}
    \caption{Architecture of the HARMONI modules integrated in the robotic coach system. }
    \label{fig:module}
\end{figure}

\subsection{VITA Modules}

We designed the robotic coach system architecture by including sensor, detector, actuator, dialogue and decision modules using the open-source framework  HARMONI \cite{spitale2021composing}. 
Figure \ref{fig:module} represents the modules integrated in the robotic coach via the HARMONI framework. We describe each of these modules in the following sections.

\subsubsection{Detector modules}
Besides relying on the robot's sensors (microphone and camera), we embedded in the robot the following detector modules to perceive the environment and the behavioral cues of the coachee.

\begin{itemize}
    \item Voice activity detector: we used the VAD library\footnote{https://github.com/marsbroshok/VAD-python} to detect when the coachee is speaking using the microphone input;
    \item Facial Action Unit (AU) detector: we integrated the OpenFace\footnote{https://github.com/TadasBaltrusaitis/OpenFace} library for real-time detection of facial action unit features;
    \item Speech feature detector: we integrated openSMILE\footnote{https://www.audeering.com/research/opensmile/} library for real-time  extraction of speech features, e.g., pitch;
    \item Facial expression recognition detector: we integrated the FaceChannel library\footnote{https://github.com/pablovin/FaceChannel} to extract the valence and arousal from coachees' facial gestures;
    \item Interaction rupture detector: we embedded the interaction rupture detector (described in Section 3.3);
    \item Speech to text module: we used the Google Speech-to-text API\footnote{https://cloud.google.com/speech-to-text} to transcribe the speech of the coachees.    
\end{itemize}

\subsubsection{Actuator modules}
Besides the wrapper for the robot motor, we embedded the robot with actuators to make the interaction more natural:.
\begin{itemize}
    \item Facial expression actuator: we used the CordialFace\cite{short2016towards} to control the facial expressions of the robot and display the lip-sync on the face-screen;
    \item Text-to-speech actuator: we used Amazon Polly Text-to-speech API\footnote{https://aws.amazon.com/polly/} to synthesize the speech of the robot. We picked Amy voice as in \cite{spitale2023robotic}.
\end{itemize}

\subsubsection{Dialogue modules} 
We implemented the following dialogue modules:

\begin{itemize}
    \item Natural language processing module: we used chatGPT to analyse the coachees' responses by integrating the OpenAI API\footnote{https://platform.openai.com/} in the robotic system;
    \item ADAPT-RL: we integrated the trained Adapt-RL model reported in Section 3.2 to make sequential decisions on which action to take next for delivering the positive psychology exercises.
\end{itemize} 

\subsubsection{Decision manager module} 
We implemented an high-level decision module as a behavior tree by using the py-tree library\footnote{https://py-trees.readthedocs.io/en/devel/} to manage the whole interaction.

\subsection{Dialogue Flow of Adaptive Coaching}
\label{sec:adapt}

In previous studies \cite{spitale2023robotic, axelsson2023robotic},  participants noted that the robotic coach was not interactive and responsive enough due to the pre-scripted nature of the interaction (\textbf{L1}). Also, coachees revealed that they would expect the robotic coach to adapt and personalise more (\textbf{L2}). VITA attempts to overcome these limitations by providing a system that enables fully autonomous and adaptive robotic coaching to improve the perceptions of the coachees towards the robotic coach (\textbf{O1}) and emulate the behaviour of a human coach (\textbf{O2}). To this end, a robotic coach needs to be trained with human coach data. Hence, we collected a dataset of human-human dyadic interactions during positive psychology  practices as existing literature lacks such a dataset.

\subsubsection{HHI4PP Dataset}
\label{sec:data}

We collected a dataset of human-human dyadic interactions between a human well-being coach and five participants (coachees), named Human-Human dyadic Interactions For Positive Psychology practice (HHI4PP). The coachees were 3 females and 2 males 29-33 years old (M=30.6 y.o., SD= 2.19 y.o.) who work as researchers at the Affective Intelligence and Robotics Lab, University of Cambridge. The human well-being coach delivered four different positive psychology exercises (about 15 minutes each). The exercises were the same as in  \cite{spitale2023robotic} and \cite{spitale_axelsson_kara_gunes_2023} -- namely, focusing on savouring, gratitude and accomplishment -- except for the fourth exercise (i.e., optimism about the future) that was substituted with the one-door closes one-door opens exercise -- which still focuses on optimism -- because participants in \cite{spitale2023robotic} found the previous exercise particularly difficult (as reported in \cite{spitale_axelsson_kara_gunes_2023}). The audio-visual recordings of the sessions were done via MS Teams, and we transcribed the speech using the automatic MS Teams transcriber. Our dataset includes a total of 5.5 hours of video recordings with the corresponding transcriptions of coach-coachee dyadic interactions. 
The ethics committee of the University of Cambridge approved the collection of this dataset and the participants signed an informed consent before taking part in this study.

\subsubsection{Reinforcement Learning Problem Formulation}
From the HHI4PP dataset,we observed (by sorting through the videos and labelling the actions) that the human well-being coach adopted a specific conversational flow to deliver the positive psychology practice, namely after first asking about an experience or a memory recall (e.g., a savouring memory) they decided whether to ask a follow-up question, summarize what the coachee just shared, or ask for a new episode (i.e., a new experience or memory). 
One of the main limitations of previous works \cite{spitale2023robotic, jeong2020robotic} was the pre-scripted nature of the coaching dialogue flow that made the interaction very repetitive and not responsive.
Hence, we formulated the problem of conducting the well-being practice in HRI as a sequential decision-making problem for selecting the next action in the \textit{coaching dialogue flow}. 
At every turn $t$ (i.e., the robotic coach asks questions, and the participant answers) \cite{gillet2022learning}, the robotic coach environment is captured as an observation state variable $s_t$. The robotic coach, similar to the human well-being coach, may choose an action $a_t$ at every turn $t$ that allows it to move forward with the conversation for delivering coaching practice by asking for a follow-up question, summarising what the coachee just said, or asking for a new episode. 
The main goal is to learn a conversational policy $\pi$: $s_t$ $\rightarrow$ $a_t$ that enables the successful delivery of the well-being practice (i.e., the robotic coach can deliver successfully the positive psychology exercise to promote coachee mental well-being). 
We formulated this as a batch (or \textit{generic}) reinforcement learning (RL) problem -- i.e., the RL model was pre-trained by using the dataset described in Section \ref{sec:data} -- and \textit{adaptive} RL -- where the generic pre-trained RL model kept learning during the interaction with the coachees. 
We used the HHI4PP dataset because it includes audio-visual recordings and transcriptions of a human coach delivering a positive psychology practice, from which our model could learn how to make a decision on the coaching dialogue flow.

We created an 11-element vector, called $s_t$, to represent the \textbf{observation state} for the coaching dialogue flow, which consists of: prediction of interaction rupture (present or absent), current well-being exercise (savouring, gratitude, accomplishment, one door closes one door opens), speech features (duration of speech and silence), and previous actions (summarisation, follow-up question, and new episode). 
All of these features were collected at the end of each turn $t$ to keep track of the dialogue flow and the conversational interchange between the human coachee and the robotic coach, as in \cite{zhang2023social}. 
The actions $a_t$ were 3 discrete dialogue \textbf{actions} of the robotic coach that can decide the coaching dialogue flow of the well-being practice, namely (1) summarise what the coachee said, (2) ask for a follow-up question (e.g., "How does this event make you feel?"), and (3) ask to share another episode (e.g., "Can you share with me another thing you were grateful for during the last week?").
We defined the \textbf{reward} as the behavioural cues of the coachees, specifically as the combination of speech duration and facial valence. We decided to use the facial valence as a component of the reward since it carries information about the emotional state of the coachee \cite{adolph2010valence} as also was done in a previous work \cite{mcquillin2022learning}, while we used the speech duration because we observed from the HHI4PP dataset that the human coach decided the coaching flow according to the amount of information shared by the coachee.  
We computed the facial valence similarly to \cite{mcquillin2022learning}, using deviations in valence values determined from coachees’ facial expressions compared to their baseline facial expression (we refer to this as $FV_t$).



We followed the same procedure for the speech duration by normalizing the values of speech duration ($SD_t$). 
The final reward was the combination of these two values as follows:
\begin{equation}
    R[s_t] = FV_t + SD_t
\end{equation}

The distribution of the reward of the HHI4PP is characterised by a mean value equal to \textit{-2.72}, a standard deviation equal to \textit{5.59}, and a median value equal to \textit{-3.09}. The total number of samples is \textit{748}.

\begin{figure}[htb!]
    \centering
    \includegraphics[width=0.95\columnwidth]{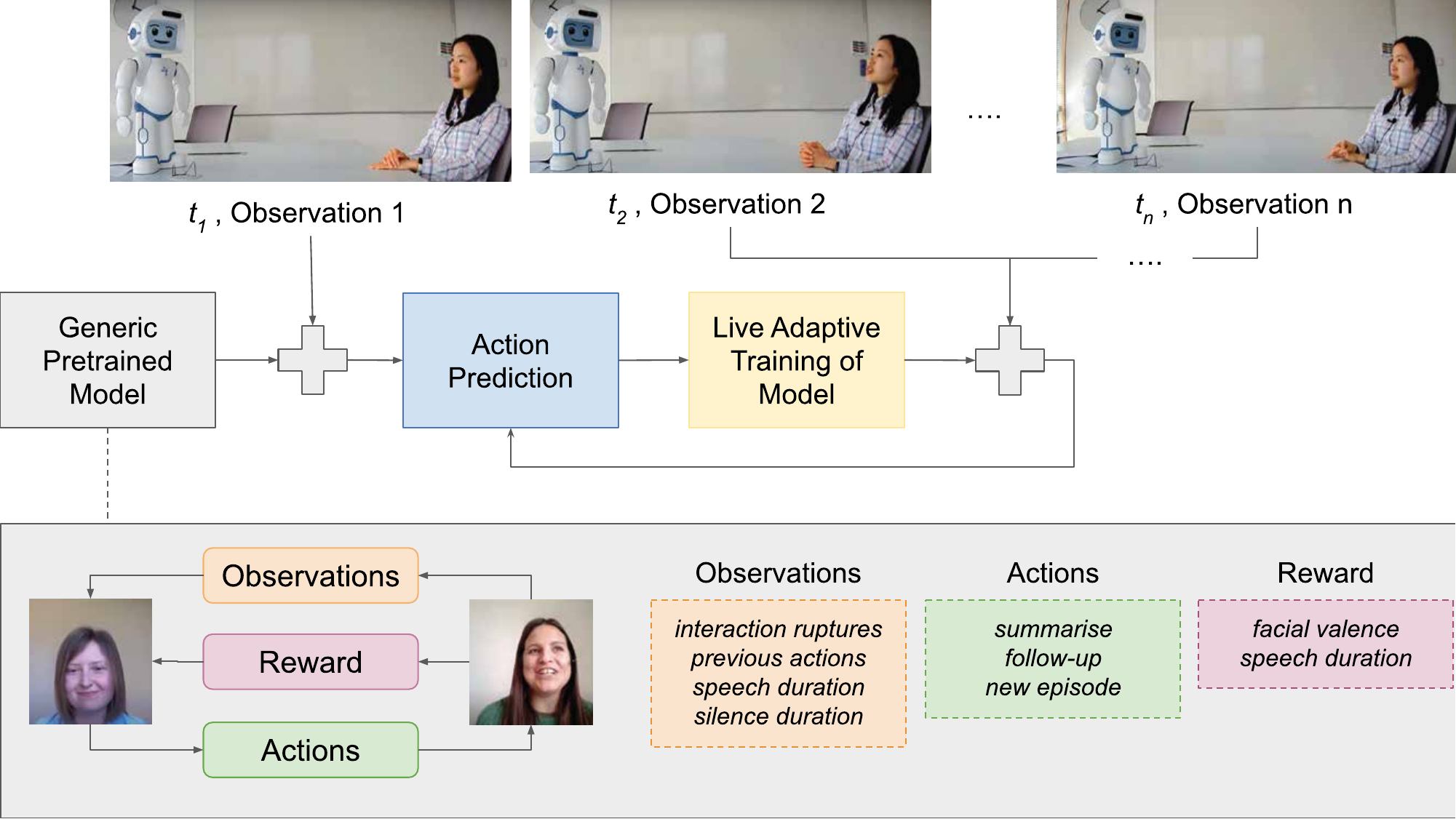}
    \caption{Reinforcement learning models: generic pre-trained model (HHI4PP dataset), and the adaptive reinforcement learning  model fine-tuned with the real-time coachee data.}
    \label{fig:system-pipeline}
\end{figure}

\subsubsection{Reinforcement Learning Models}
\label{sec:offrl}

We trained our batch (\textit{generic}) RL model using three different approaches: Deep Q-Learning (DQN), Double Deep Q-Learning (D-DQN), and Neural Fitted Q Iteration algorithm (NFQ) to learn the conversational flow of a positive psychology exercise, using the d3rlpy library\footnote{\url{https://d3rlpy.readthedocs.io/}}. 
We opted for deep Q-learning models due to their effectiveness in handling increased complexity in the number of states and actions. These models efficiently adapt to the randomness of the environment (as in-the-wild human-robot interaction), outperforming simpler Q-learning models in this regard \cite{fan2020theoretical}.
The results showed that there were not significant differences among the three models. We then chose the DQN as it is the most commonly used one in discrete problems, e.g., \cite{zhang2023social}. 


Our \textit{adaptive} RL model uses the pre-trained \textit{generic} RL model and is designed to be fine-tuned with real-time data of each coachee interacting with the robotic coach as depicted in Figure \ref{fig:system-pipeline}. We created personalised models (i.e., one for each coachee) that could further learn and adapt to the specific coachee behaviour over time (i.e., across multiple interactions).
We picked the DQN model to pre-train our \textit{adaptive} RL model, and to learn in real-time using the same d3rlpy library. This adaptive approach differs from the generic one because it keeps learning online from the data during the interaction with each coachee as in Figure \ref{fig:system-pipeline}.

\subsection{Automatic Detection of Interaction Ruptures}
\label{sec:ir}

\begin{figure}[htb!]
    \centering
    \includegraphics[width=\columnwidth]{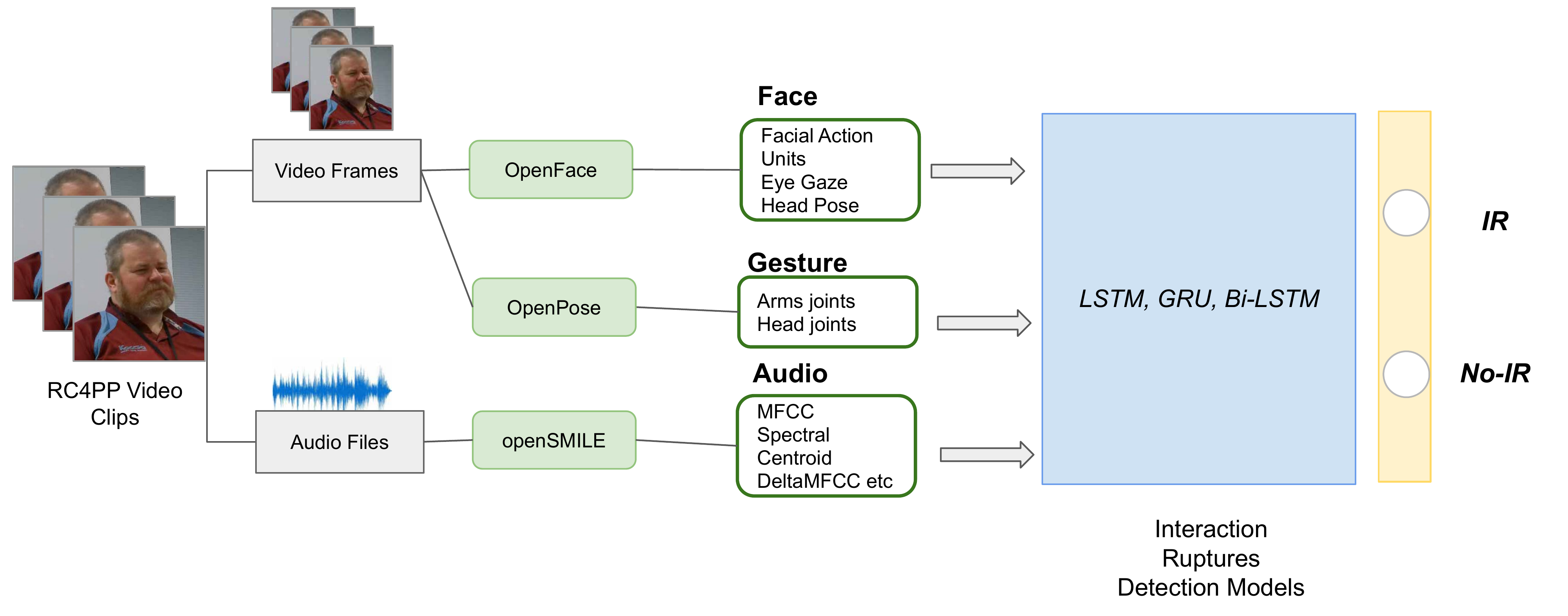}
    \caption{The model pipeline for detecting interaction ruptures using the dataset collected in \cite{spitale_axelsson_kara_gunes_2023}.}
    \label{fig:ir-model}
\end{figure}

Findings from a previous work \cite{spitale_axelsson_kara_gunes_2023} showed that several interaction ruptures may occur during a coachee-coach interaction (\textbf{L3}). 
Motivated by these findings, we decided to embed the robotic coach with the capability to detect \textit{interaction ruptures} (i.e., the user expression of awkwardness and/or reactions to the robot mistakes \cite{spitale_axelsson_kara_gunes_2023}) (\textbf{O3}).
The term \textit{interaction ruptures} refers to the  “deterioration in the quality of the relationship between patient and coach” \cite{safran1996resolution} that can occur when, for example, the robot makes mistakes during the coaching session which might negatively impact the alliance and trust perceived by the user towards the robot, which may lead to an unsuccessful well-being practice \cite{spitale_axelsson_kara_gunes_2023}.

\subsubsection{Dataset \& Labels}
In \cite{spitale2023robotic}, a robotic positive psychology coach has been deployed at a workplace for over four weeks. 
The robotic coach followed a pre-scripted interaction and it conducted four positive psychology exercises, described in detail in \cite{spitale2023robotic}. 
We requested the collected dataset from \cite{spitale_axelsson_kara_gunes_2023} which contains the extracted audio-visual features of 100 videos of about 10 minutes each that were annotated in terms of occurrences of interaction ruptures with binary labels (i.e., 1: present, or 0: absent), as described in \cite{spitale_axelsson_kara_gunes_2023}. 
The dataset \cite{spitale_axelsson_kara_gunes_2023} includes for each video clip:
    (i) 35 temporal facial features extracted using the OpenFace 2.2.0 toolkit \cite{openface2018}, such as AU1 (inner brow raiser), AU2 (outer brow raiser), AU4 (brow lowerer), AU5 (upper lid raiser) etc.; and 
    (ii) 25 temporal interpretable speech features using the openSMILE toolbox \cite{eyben2010opensmile}, such as loudness, pitch,  length of the coachees' silence and speech etc.
The dataset also includes body features, but we discarded them since they have been found to not contribute to the detection of interaction ruptures 
\cite{spitale_axelsson_kara_gunes_2023}. 

\subsubsection{Feature Pre-processing}

We used the sequences of both audio and visual features with the purpose of learning temporal patterns in audio and facial cues to predict \textit{interaction ruptures}. To this end, we created the input samples for our models by re-sampling all feature sequences at intervals of 1 second to capture the temporal variability of the data. We then applied to both feature sets a sliding window approach with a window size of $10s$ (chosen after running a set of experiments with $5$s, $10$s, $15$s, $20$s time window) and an overlap of $3s$ between sequential windows. 
Given the unbalanced nature of the dataset \cite{spitale_axelsson_kara_gunes_2023}, we decided to under-sample our feature sets using the NearMiss method that selected examples based on the distance of majority class examples to minority class examples. Our final tensors have the following dimensions (number of samples, window size, features): audio feature set (3602, 10, 25) and facial feature set (3602, 10, 35). 

\subsubsection{Automatic Detection}

We formulated the \textit{interaction rupture} (IR) detection as a sliding-window sequence-based binary classification problem, to detect whether an interaction rupture was present (IR) or not (No-IR) in a 10-seconds time window of an audio-visual segment (refer to Figure \ref{fig:ir-model} for the model pipeline). 
We experimented with three temporal deep learning models: long short-term memory networks (LSTM) \cite{hochreiter1997long}, gate recurrent unit networks (GRU) \cite{dey2017gate}, and bi-directional LSTM networks (Bi-LSTM).
We conducted a 5-fold stratified cross-validation, repeated 10 times (50 folds in total) with a subject-independent approach (i.e., the samples of the same subject were included either in the training or test set but not both). 
We chose stratified cross-validation to maintain the same proportion of “IR” and “No-IR” participants in each fold. We also tuned the hyperparameters on the training sets with the Optuna framework \cite{akiba2019optuna}.
Within each cross-validation experiment, all features in the training and testing set of each fold were standardized using the feature distributions of the training set via z-normalization, as in \cite{mathur2021modeling}.

We conducted multiple experiments to compare the uni-modal vs. multi-modal  approaches as in \cite{cheong2023s}. We first trained our models using either facial or audio features, and then we adopted two fusion strategies (either early or late fusion) to experiment with the multi-modal approaches, i.e., including audio and facial features as in  \cite{cheong2023s}.  The early fusion strategy consisted of concatenating features from the audio and facial modalities that resulted in a single vector of features. While the late fusion strategy consisted of majority voting, namely the final decision is made according to the classifier whose predicted class probability is the highest across the different uni-modal models \cite{cheong2023s}. 



\begin{table}[htb!]
    \centering
    \begin{tabular}{ccccc}
    \hline
    
    \textbf{Model} & \textbf{ACC} & \textbf{Precision} & \textbf{Recall} & \textbf{F1} \\
    \hline
    \multicolumn{5}{l}{\textbf{Facial}} \\
    \hline
    
     LSTM   & 0.58$\pm$	0.10	&0.62$\pm$	0.08&	0.58$\pm$	0.10	&0.56	$\pm$0.11  \\
     Bi-LSTM  & 0.59$\pm$	0.10	&0.63$\pm$	0.10&	0.59$\pm$	0.10	&0.57	$\pm$0.10 \\
     GRU &0.58	$\pm$0.10	&0.63$\pm$	0.09	&0.58	$\pm$0.10	&0.57	$\pm$0.11 \\
     \hline
     \multicolumn{5}{l}{\textbf{Audio}} \\
    \hline
    LSTM  &  \textbf{0.73 $\pm$	0.09}&	0.77	$\pm$0.07&	\textbf{0.73	$\pm$0.09}&	\textbf{0.73$\pm$	0.09}  \\
     Bi-LSTM &  \textbf{0.73$\pm$	0.10}&	0.78$\pm$	0.08	&\textbf{0.73$\pm$	0.10}&	\textbf{0.73$\pm$	0.10 }\\
     GRU & 0.70 $\pm$ 0.10&	0.77$\pm$0.08	& 0.72$\pm$0.10 &0.72$\pm$0.09 \\
     \hline
     \multicolumn{5}{l}{\textbf{Facial + Audio (Early Fusion)}} \\
    \hline
     LSTM  & 0.54$\pm$	0.09	&0.52$\pm$	0.10	&0.52$\pm$	0.09	&0.51$\pm$	0.11  \\
     Bi-LSTM  & 0.55$\pm$	0.09	&0.58$\pm$	0.10	&0.55$\pm$	0.09	&0.52$\pm$	0.11 \\
      GRU &0.55$\pm$	0.09	&0.52$\pm$	0.10	&0.53$\pm$	0.09	&0.51$\pm$	0.11\\
     \hline
     \multicolumn{5}{l}{\textbf{Facial + Audio (Late Fusion)}} \\
    \hline
     LSTM  & 0.67$\pm$	0.10	& 0.80$\pm$0.09	&0.53$\pm$	0.18&	0.62$\pm$	0.16 \\
     Bi-LSTM &  0.68$\pm$	0.12	&\textbf{0.81}$\pm$\textbf{	0.16}	&0.53$\pm$	0.19&	0.63$\pm$	0.17 \\
     GRU & 0.68$\pm$	0.11	&0.78$\pm$ 0.15	&0.52$\pm$	0.10&	0.63$\pm$	0.18 \\
     \hline
    \end{tabular}
    \caption{Experimental results for uni-modal and multi-modal approaches using facial and audio features across three deep learning models (LSTM, Bi-LSTM, and GRU) averaged across the 50 folds. }
    \label{tab:ir-results}
\end{table} 

\subsubsection{Evaluation Metrics}

The evaluation metrics computed for each of the 50 cross-validation folds included: (1) accuracy (ACC), the number of correct predictions with respect to the total number of predictions in the test set; (2) precision, the percentage of interaction rupture ("IR") samples among all samples that were classified as ruptures; (3) recall, the percentage of samples accurately identified as interaction ruptures ("IR") among all "IR" samples in the dataset; and (4) F1, the combination of the precision and recall of the classifier (by taking their harmonic mean), as in most classification problems \cite{hossin2015review}.

These metrics were inspected across cross-validation folds and used to examine the performances of various modelling strategies. Our problem is an event detection task similar to \cite{kontogiorgos2020behavioural} where the model should detect interaction ruptures using a sliding-window approach in an HRI context. In this scenario, as in \cite{mathur2021modeling}, the precision of the interaction rupture detection is crucial (e.g., a robot relying on precise interaction rupture detection to appropriately repair those ruptures) to minimise false positives and maintain coaching effectiveness. 
For this reason, our primary metric for identifying the highest-performing model was \textit{precision}.
We prioritised precision rather than the precision-recall tradeoff. Discussions with a human coach led to our decision that predicting false interaction ruptures and making repairs based on this can be very costly in a coaching practice.

\subsubsection{Results}

Table \ref{tab:ir-results} reports the results obtained from the experiments for uni-modal and multi-modal modeling of LSTM, GRU, and Bi-LSTM models.
Our results show that the Bi-LSTM late fusion approach achieved an average  precision of 81\%, substantially higher than the average precision of other models (around 66\%). Given that precision is our primary metric, the late fusion multi-modal Bi-LSTM approach has been chosen as the model to embed in our robotic coaching system.

\subsection{Prompt Engineering of LLM}
\begin{table}[htb!]
    \centering
    \footnotesize
    \begin{tabularx}{\columnwidth}{s|X}
    \toprule
    \textbf{Action} &  \textbf{Prompt Engineering}\\
        \midrule
        (1) Summarising &Can you please summarise what the Human has just shared?  \\
        \midrule
        (2) Asking for a follow-up question & Can you please ask me a follow-up question about the exercise episode I have just shared?  \\
        \midrule
        (3) New episode & Can you please ask me about a new episode to share? \\
    \bottomrule
    \end{tabularx}
    \caption{Prompt engineering for making a completion request to ChatGPT in a coaching practice. }
    \label{tab:prompt}
\end{table}

\label{sec:prompt-engineering}

To achieve a more natural and multi-modal interaction and embed the robotic coach with more sophisticated conversational capabilities (\textbf{L4}), the robotic coach has been designed to analyse the coachees' speech content and use a Large Language Model (LLM, i.e., ChatGPT) to understand what the coachee says and respond appropriately to imitate a human coaching practice (\textbf{O4}). We have used the LLM with all the precautions needed by implementing several safety measures to avoid any unexpected or inappropriate responses as recommended in \cite{axelsson2022robots} and \cite{cabrera2023ethical}. First, we integrated a natural language processing layer (i.e., Moderation APIs from OpenAI) to identify whether the coachee’s request content is inappropriate (e.g., self-harm, sexual) and whether the LLM’s answers are inappropriate. If the layer detected an inappropriate request, the robotic coach would answer "I found your answer very inappropriate. I would stop here the coaching practice and call the researcher", and stop the interaction. Second, we conducted several adversarial testing to better understand how the LLM replies to controversial requests (e.g., the coachee shares that they have punched someone at work), see Section \ref{app:adv-test} of the Appendix for more examples. Finally, we have implemented “Prompt engineering” to constrain the topic and tone of the output text. This aims to reduce the possibility of producing undesired content, even if a coachee would try to produce it.
Table \ref{tab:prompt}\footnote{Note that gpt-3.5-turbo takes as input "Human" and "AI" conversational utterances.} summarises the prompts for each action of the robotic coach. 
Note that the inclusion of the Moderation API layer caused a lag in the response of the robot, but we thought it was essential for a safe and ethical conversation with the robotic coach.

We have used \textit{gpt-3.5-turbo} model to request to "complete" the conversation between the coachee and the robotic coach. We fed the model with the system context (i.e., the description of the exercise that the robotic coach has to deliver), and then we incorporated the coachee's transcriptions. We repeated the same prompting (appending one of the prompts reported in Table \ref{tab:prompt} according to the decision made by the robotic coach as described in Section \ref{sec:adapt}) for the whole coaching practice. We kept also the history of the dialogue in the memory of the LLM in order to provide the LLM model with the context of the conversational coaching practice. An example of the conversation is reported in Section \ref{app:ex} of the Appendix.

The ultimate goal of VITA is to understand whether it can enable a significant improvement of mental well-being not only limited to short-term interactions (\textbf{L5}) but also in the long-term and in real-world settings (\textbf{O5}).

\subsubsection{Considerations for using LLMs}
\label{sec:ethical-lessons}

Using LLMs to generate content for a robotic coach requires several important considerations prior to and during an empirical human-robot interaction study. We share below our experiences and recommendations in order to guide the researchers that intend to use the VITA system in their own research works.

Firstly, prior to the study, a thorough ethics review of the empirical study was undertaken by our departmental ethics review committee at the Department of Computer Science and Technology, University of Cambridge. They reviewed and approved the study design, the experiment protocol, and the consent forms. We recommend that researchers considering using the VITA system should engage with the appropriate ethics committee at their institution, and discuss with them whether its use is appropriate and what safety measures should be taken. 

Secondly, we implemented several safety measures withing the system while using ChatGPT (as detailed in Sec. \ref{sec:prompt-engineering}): adding an NLP layer to detect inappropriate user generated language, conducting adversarial testing for controversial requests (see Appendix \ref{app:adv-test}), and ``prompt engineering'' to constrain the topic and tone of the output. These measures were taken to keep the well-being interaction appropriate and on topic. We recommend that researchers using the VITA system also undertake such careful measures, and particularly consider what kinds of adversarial tests are relevant for their intended use and context, and which prompts appropriately guide and constrain the intended interaction.

Thirdly, to further ensure that the interaction was kept within the boundaries of an appropriate well-being interaction session, we scaffolded the aforementioned engineered prompts within a structured well-being interaction with pre-scripted robot utterances (see Appendix \ref{app:ex} for details). 
This decision aligns with our previous work on design and ethical recommendations for robotic well-being coaches \cite{axelsson2022robots}, where professional human practitioners recommended that ``verbal adaptation should be limited to preserve well-being practice efficacy''. By scaffolding LLM-generated (i.e., verbally adapted) sentences within a structured interaction, we aimed to apply the appropriate level of adaptation for this well-being interaction, while also reducing the possibility of the interaction extending outside the context of a well-being interaction, and minimizing the possibility of misunderstandings via inappropriately generated utterances. We recommend that researchers using the VITA system consider what level of verbal adaptation is appropriate and safe for the intended interaction, and apply structured scaffolding as appropriate.  

Finally, we recruited a study population that we pre-screened for anxiety and depression (as discussed in Sec. \ref{sec:participants}). We chose this approach in order to safeguard our participants prior to the interaction (as also recommended in our prior work \cite{axelsson2022robots}). We consider it ethical to thoroughly examine and investigate well-being technologies with populations without mental health difficulties, in order to mitigate potential risks that may arise with vulnerable populations. We recommend that researchers evaluate how appropriate using the VITA system is for their intended application context, and whether their population  includes users who may be vulnerable to potential discomfort arising in the interaction, or who might need additional safety measures.

We do urge researchers using the VITA system in the future to carefully consider all of these recommendations. However, this is not to say that merely following or considering these ethical recommendations is sufficient for an approach, interaction or application area to be ``ethical''. There is limited prior work on the ethics of implementing LLM models in embodied systems, particularly for well-being. Future research is needed to thoroughly investigate these issues --- especially prior to applying such technology outside of a controlled and safely monitored research environment.





\section{Pilot Study}
\label{sec:pilot}
This section describes the in-the-lab pilot study to evaluate the use of different configurations of the robotic coach interacting with coachees during the coaching practice to address \textbf{O1} and \textbf{O2}.

\subsection{Participants \& Protocol}
We collected a dataset of 4 participants (2 males, and 2 females, 20-33 years old) from the Department of Computer Science and Technology, University of Cambridge, interacting with three configurations of the robotic coach delivering four positive psychology exercises: \textit{pre-scripted} interaction as in \cite{spitale2023robotic} (PS, i.e., the robotic coach follows a pre-defined sequence of utterances to speak aloud to the coachee no matter what the coachee says), \textit{generic} RL (GEN-RL, i.e., the robotic coach understands via natural language processing what the coachee says and responds accordingly without personalising the selection of the next line of the dialogue to each coachee’s behaviour), and \textit{adaptive} RL (ADAPT-RL, i.e., the robotic coach understands via natural language processing what the coachee says and personalises to the coachee’s behaviour by selecting the next line of the dialogue flow accordingly), see Section \ref{sec:offrl}. 
The robotic coach delivered the same four exercises (savouring, gratitude, accomplishment, and one door closes one door opens) of the HHI4PP dataset described in Section \ref{sec:data}.  
Each participant interacted with 2 of the 3 robotic coach configurations (either PS and GEN-RL or PS and ADAPT-RL). We recorded audio-visual clips via an external webcam and we gathered a total of 6 hours data in 32 HRI sessions.

\subsection{Robotic Platform}
We chose the QTrobot by LuxAI S.p.A as the robotic platform to deliver the PP exercises given its flexibility (ROS-based) to implement the VITA system as in previous works \cite{spitale2023robotic, axelsson2023robotic, spitale2022socially}. As pinpointed by \citet{spitale2022socially}, the QTrobot has the advantage to be equipped with a screen face, a RGB-D camera, and a microphone array that can be used for real-time computational analysis.
For instance, \citet{spitale2023robotic} embedded the robotic coach with less sophisticated skills that matched more with the Misty Robot’s form.  However, we hypothesised that by equipping the QT robot (a humanoid robot) with more advanced capabilities, provided by VITA, it would better match with the skills and behaviours exhibited while delivering the coaching practice.
The QTrobot is a 90 cm tall, child-like robot, with 2 DOF neck, 4 DOF full arms (shoulders, elbows, and hands), and static legs.
We implemented the VITA robotic coach system described in Section \ref{sec:system} on the QTrobot. 

\subsection{Measures}
During the interaction, we gathered the \textbf{logs} of the interaction (e.g., transcriptions of the coachees' speech, responses of the robot etc.) and also the IR and RL model outputs (i.e., IR predictions, actions chosen by the robot, rewards) to evaluate and analyse the VITA components. 
At the end of each robotic coach interaction, we asked participants to fill out (once for each condition they experienced) two \textbf{questionnaires}, namely the  Negative Attitude Towards Robots Scale (NARS) \cite{nomura2006measurement}, and the Game Experience Questionnaire (GEQ) \cite{ijsselsteijn2013game}, with general, flow, and tension sub-scales to evaluate the coachees experience interacting with the robot as in  \cite{mcquillin2022learning}, and overall impression, which included customised questions, such as "I felt that the robot's behaviour was adapting to what I was doing", "I felt that the robot speech was adapting to what I was saying" etc.

 \begin{figure}[htb!]
     \centering
     \begin{subfigure}[b]{0.48\columnwidth}
         \centering
         \includegraphics[width=\textwidth]{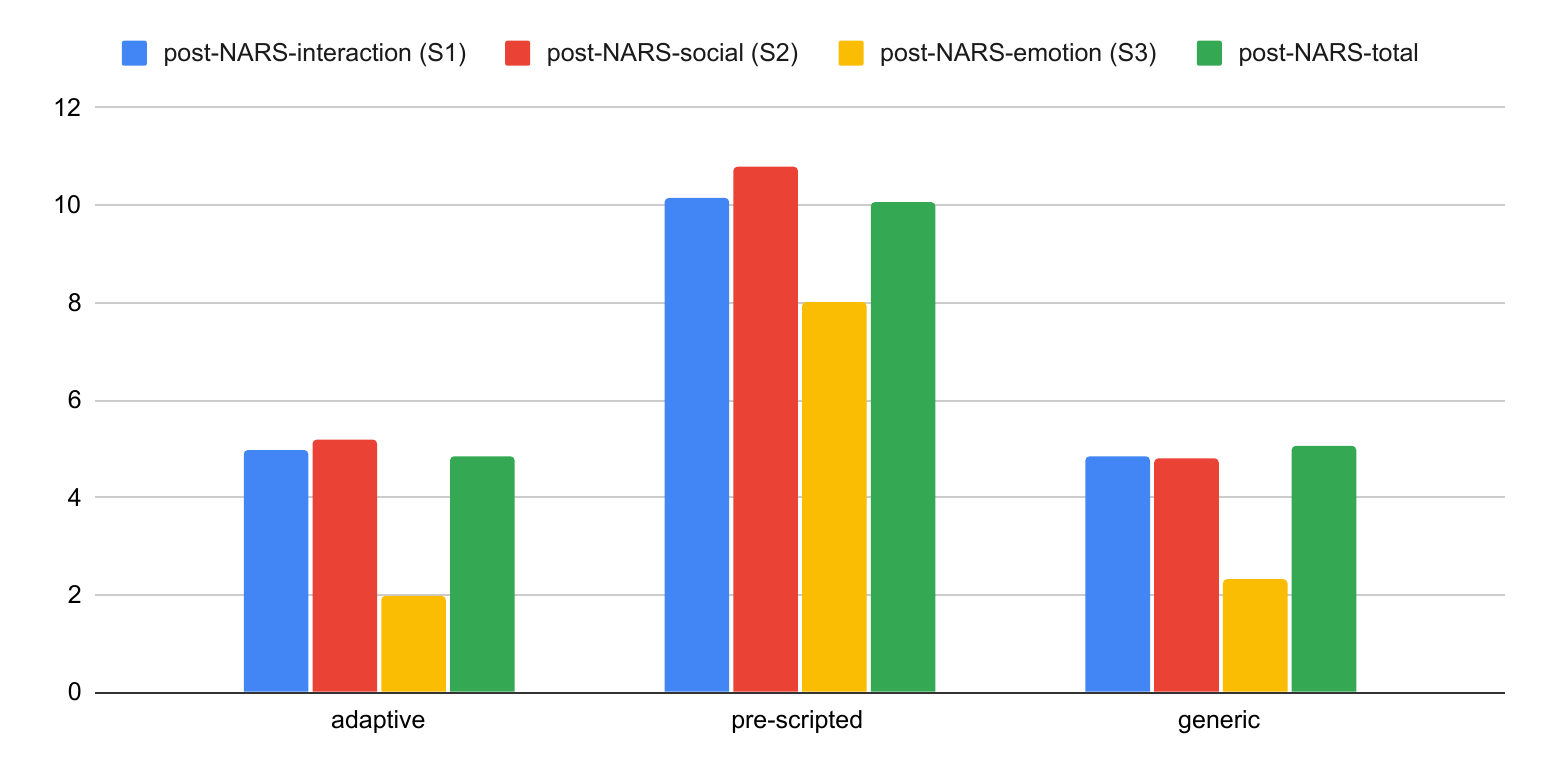}
         \caption{NARS evaluation post-study for the pre-scripted (PS), adaptive (ADAPT-RL), and generic (GEN-RL) configurations.}
         \label{fig:nars-pilot}
     \end{subfigure}
     \hfill
     \begin{subfigure}[b]{0.48\columnwidth}
         \centering
         \includegraphics[width=\textwidth]{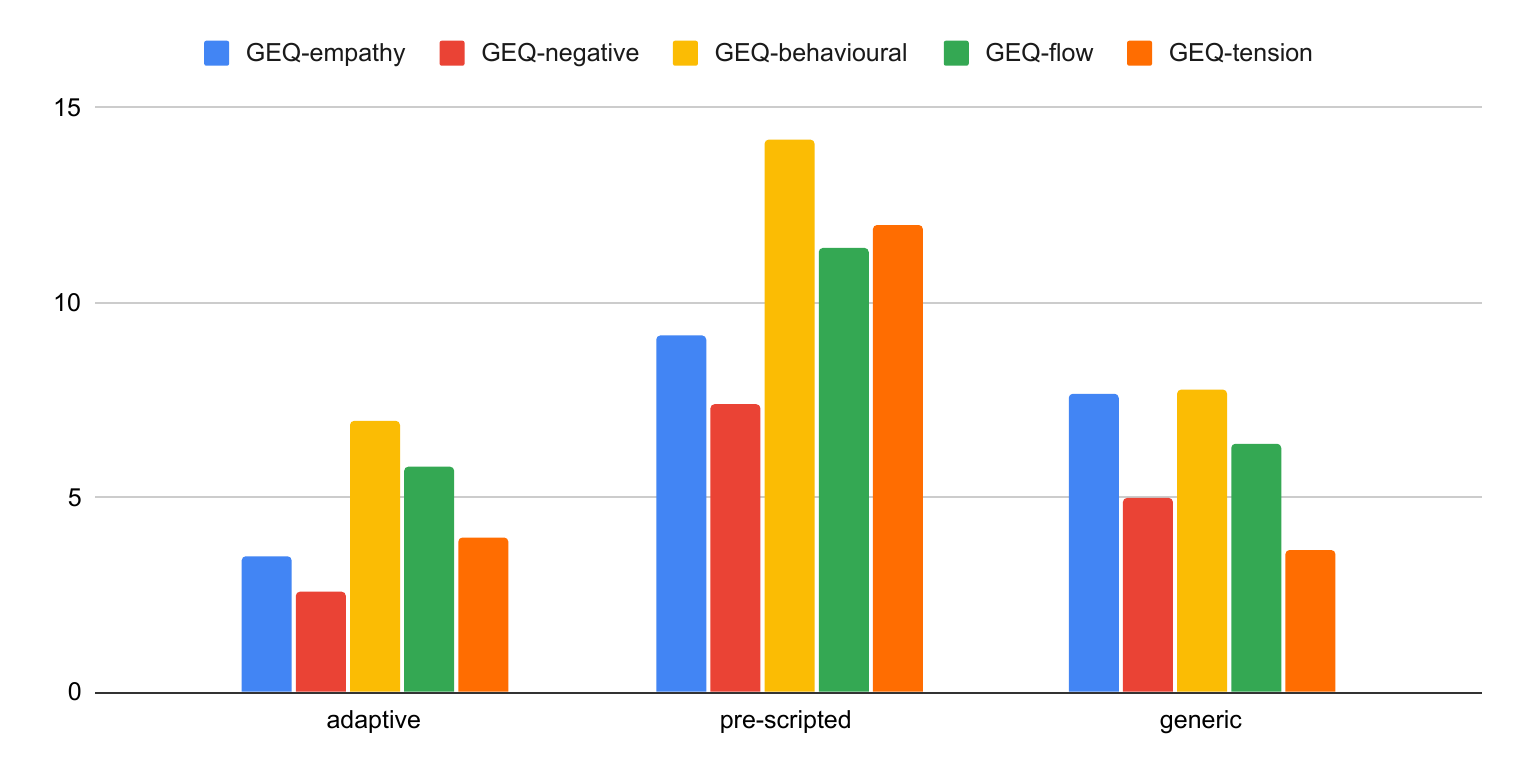}
         \caption{GEQ evaluation post-study for the pre-scripted (PS), adaptive (ADAPT-RL), and generic (GEN-RL) configurations.}
         \label{fig:geq-pilot}
     \end{subfigure}
     \hfill
        \caption{Questionnaire results from the pilot study.}
        \label{fig:pilot_results}
\end{figure}

\subsection{Results}
\label{sec:pilot-results}
We conducted the exploratory pilot study with a small number of participants to inform the design of our real-world study, and as such we cannot conduct statistical analysis on the results. 
Instead, we compared the perception of the coachees between the PS vs. GEN-RL and ADAPT-RL configurations (see Figures \ref{fig:nars-pilot} and \ref{fig:geq-pilot}). In brief, we observed that in the PS configuration the coachees displayed the most negative attitude towards the robot (NARS), they felt the interaction flow was not very engaging (GEQ-flow), they felt tension during the coaching practice (GEQ-tension), they were not involved (GEQ-behavioral involvement) as in the other two configurations, and their overall impression of the robotic coach was very negative. 
Figure \ref{fig:nars-pilot} plots the post-study NARS results for the 3 sub-scales, namely interaction, social and emotion) in the three configurations (pre-scripted, generic and adaptive). The results suggest that the coachees have a less negative attitude towards the robotic coach in the generic and adaptive configurations with respect to the pre-scripted one.
Analogously, Figure \ref{fig:geq-pilot} depict the post-study GEQ results of the sub-scales empathy, negative behavioural, flow, and tension in the three configurations (pre-scripted, generic and adaptive). Again, the results show that coachees felt more negative and tension towards the robotic coach in the pre-scripted configuration with respect to the adaptive and generic configurations.
Our results suggest that the \textit{generic} and \textit{adaptive} configurations were perceived more positively than the \textit{pre-scripted} one (\textbf{EV1.1}, addressing \textbf{O1}). 

\begin{figure}[htb!]
    \centering
    \includegraphics[width=0.85\columnwidth]{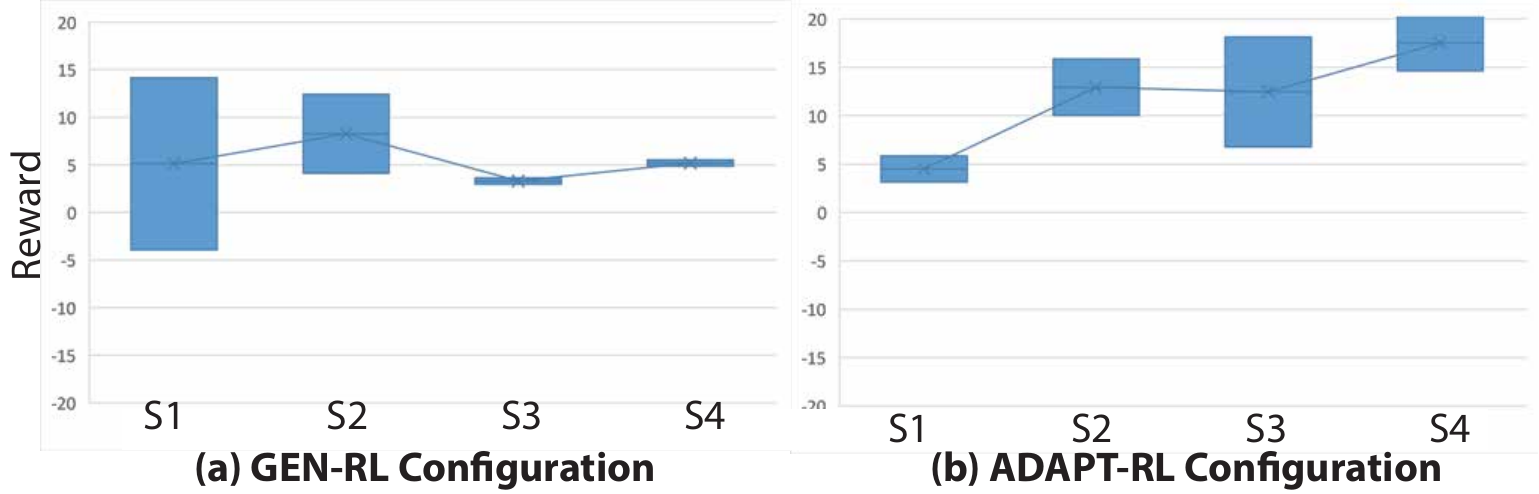}
    \caption{Comparison between the rewards of the generic configuration (a), and the adaptive configuration (b). Legend: S refers to the session.}
    \label{fig:pilot_reward}
\end{figure}

We have logged the reward over the four sessions in all the configurations tested. We observed that the trend of the reward in the adaptive configuration (see Figure \ref{fig:pilot_reward}b) is positive (i.e., it tends to increase), while the trend in the generic configuration is inconsistent (i.e., neither decreasing nor increasing,see Figure \ref{fig:pilot_reward}a). These results suggest that the model in ADAPT-RL successfully keeps learning during the coaching practice (\textbf{EV2.1}, addressing \textbf{O2}). Therefore for the real-world study we adopted the ADAPT-RL configuration for the VITA system. 

\section{Real-world Study}

This study aims to address the five objectives reported in Table \ref{tab:motivation}, using the VITA robotic coach configuration resulting from the pilot study. 
This section details the real-world study by reporting the participants' information (coachees), the protocol, the measures, the data analysis and the results obtained. 



\subsection{Participants}
\label{sec:participants}

We involved 17 participants in total, 7 females, and 10 males, 4 of whom were 18-25 years old, 6 were 26-35 years old, 4 were 36-45 years old, and 3 were 46-55 years old. The study was supported and promoted by 
the Cambridge Consultants Inc. tech company located in Cambridge , and all participants were employees in this company. 
The company advertised the study via email and participation was voluntary (with no compensation).  
Our study aimed to involve healthy participants in the first instance, therefore we screened 25 participants and recruited 17. 8 participants were excluded based on their self-reported levels of anxiety and depression, scoring more than 9 (the maximum threshold for mild anxiety disorder) in the Generalized Anxiety Disorder 7 (GAD-7) \cite{spitzer2006brief} and in Patient Health Questionnaire (PHQ-9) \cite{lowe2004measuring}. Participants had very little knowledge (on average $1.53$ on a 5-point Likert scale) of robotic technology. 
All participants provided informed consent for their participation and agreed on the usage of their data for scientific research. The study design, the experiment protocol, and the consent forms were approved by the Ethics Committee of the Department of Computer Science and Technology, University of Cambridge. 


\subsection{Measures}
\label{sec:meas}
We collected quantitative and qualitative data via VITA-based logs of the interaction, questionnaires, and interviews. 
\textbf{VITA Logs} were the same logs as reported in Section \ref{sec:pilot}.
One week prior to the study, we asked participants to fill out the following \textbf{questionnaires}: a demographic form (asking their age, gender, and previous experience with well-being practices as well as robots), 
the Negative Attitude Towards Robots Scale (NARS \cite{nomura2006measurement} to measure coachees' negative attitudes towards robots before interacting with them as in \cite{spitale2022socially}),  
and Ryff’s Psychological Well-being Scale (RPWS \cite{van2004construct}) to assess coachees' mental well-being as in \cite{jeong2020robotic}. 
At the end of each robotic coach interaction, we asked participants to fill out the same questionnaires of the pre-study as well as 
the Game Experience Questionnaire (GEQ) \cite{ijsselsteijn2013game}, with general, flow, and tension sub-scales, 
and overall impression (same questions as reported in Section \ref{sec:pilot}). 
At the of the study, we conducted a semi-structured \textbf{interview}, in which we asked several questions, such as 
``Have you felt understood and listened to?'', ``Would you recommend a colleague or a friend to use this robot?'' etc. We concluded the study by debriefing the coachees, i.e., explaining the main goal of the study, and answering their questions.

\subsection{Protocol}
In this study, we chose the ADAPT-RL configuration from the pilot study to deliver the four positive psychology exercises with the QT robot, due to the results reported in Section \ref{sec:pilot}.
Prior to the study, the coachees were asked to fill out a set of questionnaires detailed in Section \ref{sec:meas}. 
The robotic coach delivered the four positive psychology exercises described in Section \ref{sec:pilot} over four weeks in a meeting room of the Cambridge Consultants Inc. company's offices on a weekly basis. Two researchers monitored the study from another meeting room in the same building. 

In each session, one of the researchers welcomed the coachee and asked them to enter the meeting room and sit on the chair in front of the robot. 
Once the coachee was ready, the other researcher started the recordings via the two cameras, and they left the room to leave the coachee alone with the robotic coach. 
The dyadic interaction between the coachee and the robotic coach lasted for about 10 minutes and included the following steps:
(1) The robot introduced itself and described the positive psychology practice (just in the first session) and the exercise of the week (e.g., savouring exercise). 
(2) The robot asked the coachee to think about a positive memory from the last week and to share it with them.
(3) The robot listened to the coachee's response.
(4) The robot made a decision -- based on the ADAPT-RL model described in Section \ref{sec:offrl} -- on the next step (summarise, ask for a follow-up question, or start a new episode).
(5) The robot generated the action according to the decision made in (4) and listened to the coachees' response.
(6) The robot repeats steps (4) and (5) for 8 turns. We decided to fix the number of turns to 8 to ensure that the coaching practice does not last more than 10 minutes.
(7) The robot concluded the session by asking the coachee to fill out the questionnaires (described in Section \ref{sec:meas}) on a tablet. Before leaving the room, the robot thanked them and reminded coachee of the following week’s session.
At the end of the four weeks, the researchers asked the coachee to fill out the final questionnaires and conducted a final interview as detailed in Section \ref{sec:meas}.

\subsection{Data Analysis}
We gathered various \textbf{logs} of the models embedded (i.e., the reward, actions, interaction rupture values) and the interactions (i.e., facial valence, speech duration) for each episode in every session with the robotic coach. 
We then computed the average value across each session and evaluated these values' evolution over 4 weeks to understand whether the ADAPT-RL model was able to learn how to adapt to each person over time (i.e., if the model has learned, the reward value would increase over 4 weeks). Then, we conducted a non-parametric statistical analysis using Friedman and then post-hoc Wilcoxon tests to evaluate the measures' change over time. 
As in \cite{spitale2023robotic}, we analyzed the quantitative data from the pre-, and post-study \textbf{questionnaires} using Python statistical libraries. We conducted non-parametric tests because our samples do not follow a normal distribution. In particular, we used Wilcoxon signed-rank test to compare the measurements pre-, and post-study with Bonferroni correction. 
We applied the \textit{framework method} to analyse qualitative data \cite{ritchie1994analysing} collected from semi-structured  \textbf{interviews}, conducted after the final session. The \textit{framework method} consists of five key stages: 1) familiarization with the data, 2) identifying a thematic framework, 3) indexing, 4) charting, and 5) mapping and interpretation.  

\subsection{Results}
This section reports the main findings that have been obtained evaluating the results collected during the study via different data sources, namely the VITA logs and questionnaires to address the objectives listed in Table \ref{tab:motivation}.

\label{sec:real-world}
    \begin{figure*}[htb!]
    \centering
    \includegraphics[width =\textwidth]{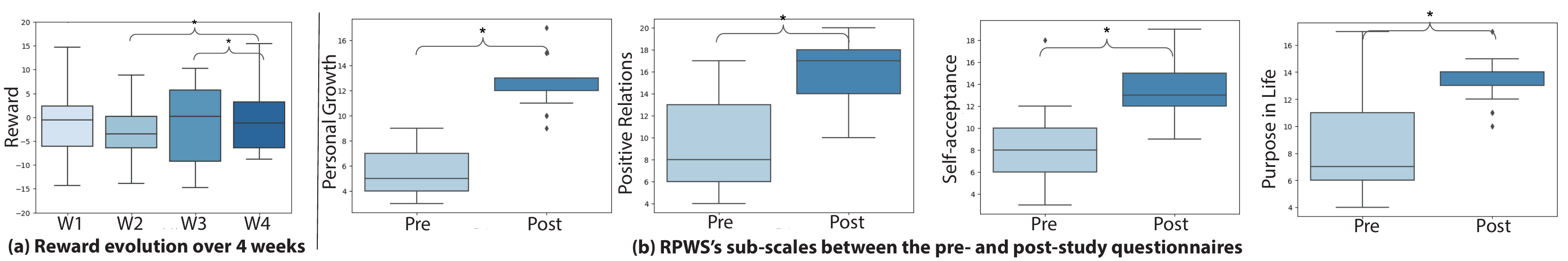}
    \caption{Results of the real-world study. Legend: W refers to the week.}
    \label{fig:realworld}
\end{figure*}

\subsubsection{VITA log results.}
\label{sec:vita-results}

We computed the rewards over 4 weeks for each participant, and then we aggregated the rewards to compute whether on average the online RL model has learned over the weeks as depicted in Figure \ref{fig:realworld}a. We then conducted a paired t-test with Bonferroni correction (0.05/4) and we found that the reward in W4 ($M = -0.06$, $SD = 8.98$) was significantly higher (W2-W4: $t = -2.42$, $p<.05$; W3-W4: $t = -2.47$ , $p<.05$) than the reward in W2 ($M = -2.80$, $SD = 7.94$) and in W3 ($M = -3.16$, $SD = 12.3$). 
Figure \ref{fig:realworld} depicts the evolution of the reward over time, and it shows that the reward increased over time from W2 to W4 except for W1. This is due to the initial model exploration in W1, as it tries to adapt the pre-trained policy towards the specific individual user (as depicted by the wide bar range). 
This result suggests that the model was able to learn on average how to adapt to the person (\textbf{EV2.2}) and contributes to addressing \textbf{O2}.
The reward increase is motivated by the speech duration change over time. 
We conducted a Friedman test for evaluating the change over time of the coachee's speech duration. The results showed that  there was a statistically significant difference in terms of speech duration variance over the four weeks ($\chi^2$ = $11.11$, $p$<$.05$).  We then conducted post-hoc tests with the Bonferroni correction (.05/4), and we found that the speech duration variation was significantly lower in W2  ($Mdn$ = $13.81$) than in W3 ($Mdn$ = $15.17$; $z=28.0$, $p$<$.05$) and W4 ($Mdn$ = $18.23$; $z=21.0$, $p$<$.01$).
The speech duration of coachees on average increased from W2 to W4. 
We analysed the number of interaction ruptures that occurred during the robotic coaching. Our results show that no interaction ruptures were detected during the coaching sessions (\textbf{EV3.1}). 
This finding is promising indicating that coachees may have not experienced any major interaction ruptures during the robotic coaching (addressing \textbf{O3}). We discuss this further in Section \ref{sec:summ}. 

\subsubsection{Questionnaire results.}
\label{sec:quest-results}

We compared the NARS sub-scales (i.e., emotion, social, and interaction) pre- and post- study using a Wilco- xon-ranked test, and our results show that the NARS emotion sub-scale score was significantly (Z = $11.50$, $p < .05$) lower in the post-study ($Mdn$ = $2.00$) than in the pre-study ($Mdn$ = $2.33$) measurement. This result suggests that the coachees had a less negative attitude towards the robotic coach after the study. 
We also observed that the GEQ questionnaire results show that coachees reported low tension ($Mdn = 1.67$) and low negative feeling ($Mdn = 1.4$) during the robotic coaching. However, they reported a positive flow in interacting with the robotic coach ($Mdn = 3$), they felt slight empathy ($Mdn = 2.67$), and they reported behavioral involvement ($Mdn = 3.67$) during the robotic coaching (\textbf{EV1.2}). These results contribute to addressing \textbf{O1}.
\begin{table}[htb!]
\footnotesize
    \centering
    \begin{tabularx}{\columnwidth}{mbmm}
    \toprule
    & \textbf{Statement} &\textbf{Mean}& \textbf{SD}\\
    \midrule
         1& I felt that the robot's behavior was adapting to what I was doing. &3.29 &1.15\\
         2&	I felt that the robot speech was adapting to what I was saying.&	4.11&0.78\\
         3& I felt that I was adapting to what the robot was doing. &3.17&1.18\\
         4& 	I felt I was adapting to what the robot was saying.&3.82&0.95 \\
         5& 	I felt that the content of the follow-up questions was appropriate. &3.88&0.78\\
         6& 	I felt that the number of the follow-up questions was appropriate. &3.88&0.85\\
         7& 	I felt that the number of the episodes that the robot asked me to share was appropriate.&3.58&1.06\\
         8& 	I felt that how the robot summarized what I said was appropriate. &3.88&0.92\\
         9&  	I felt that the number of times that the robot summarized what I said was appropriate.&3.64&1.16 \\
         \bottomrule
    \end{tabularx}
    \caption{Item-by-item overall impression of the coachees after interacting with the robotic coach over 4 weeks. }
    \label{tab:overall-impression}
\end{table}

We observed similar results for the customised questionnaire reported in Table \ref{tab:overall-impression}. Table \ref{tab:overall-impression} reports the overall impression results obtained from the customised questionnaires after interacting with the robotic coach over 4 weeks.
Overall, coachees had a positive impression of the robotic coach and they thought that it was adapting to them and it showed appropriate behaviour in terms of dialogue flow and decision-making (\textbf{EV2.3}). This result contributes to addressing \textbf{O2}.

We conducted the same type of test also for the RPWS's sub-scales (i.e., autonomy, environmental mastery, personal growth, positive relations with others, purpose in life, and self-acceptance). 
Our results show that the following sub-scales where significantly higher in the post-study than in the pre-study: the RPWS personal growth ($Z = 1.00$, $p < .001$, $Mdn_{post-study} = 13$, $Mdn_{pre-study} = 5$), the RPWS positive relations with others ($Z = 14.00$, $p < .01$, $Mdn_{post-study} = 17$, $Mdn_{pre-study} = 8$), the RPWS purpose in life ($Z = 2.50$, $p < .01$, $Mdn_{post-study} = 14$, $Mdn_{pre-study} = 7$), and the RPWS self-acceptance ($Z = 14.00$, $p < .01$, $Mdn_{post-study} = 13$, $Mdn_{pre-study} = 8$)
Figure \ref{fig:realworld}b depicts the sub-scales of personal growth, positive relations with others, purpose in life, and self-acceptance that resulted in being significantly different pre- and post- study (\textbf{EV5.1}). 
This result suggest that the coachees' self-reported mental well-being improved via the robotic coach practice and this result contributed to addressing \textbf{O5}.

\subsubsection{Interview results}
\label{sec:int-results}

Interview results help address the objectives of this study as follows. In general, coachees reported a very positive experience of interacting with the robotic coach and that they felt at ease and relaxed after the first session. For example, P10 reported that she \textit{``was more relaxed after session one because [she] didn't really know what to expect [..] relaxation happened very quickly during the first session''}. Again, P13 felt a bit \textit{``awkward''} at the beginning and then she \textit{``got kind of used to the robot''}. 
These results support \textbf{O1} (\textbf{EV1.3}).  

Coachees have contradicting opinions of their perception of the robotic coach adaptation. P02 found that the robotic coach was \textit{``adapting''} to what she said and P03 mentioned that he \textit{``thought the conversation was more natural [in the last session] than before''}. While, P11 felt that \textit{``the information about the exercises was just being read out in not a very personalized way''}. Some of the coachees felt that they were adapting to the robotic coach rather than the opposite. For example, P02 mentioned that she \textit{``adjusted the way [she] spoke''}. These results partially support \textbf{O2} (\textbf{EV2.4}).

Some of the coachees found the interaction with the robotic coach very \textit{``natural''} (P10) while others haven't found it quite seamless (e.g., P13, and P15) because of the robot's long time to respond to them and interruptions of the robot during the coaching practice. However, P06 also mentioned feeling \textit{``suprised"} because she \textit{"built up a bit the relationship''} with the robotic coach. The coachees' opinion didn't fully support \textbf{O3} (\textbf{EV3.2}). 

In general, coachees were very impressed by the conversational capabilities of the robotic coach to be able to understand the content of the coachees' speech. P16 was very \textit{``impressed with how much kind of knowledge it has and the kind of appropriateness of the follow up questions''}, and again P07 found \textit{``summarizing [of the robotic coach] fairly impressive''}. Also, coachees found that the robotic coach asked \textit{``appropriate questions''} (e.g., P08).
Most of them (e.g., P12, P13) felt \textit{``listened to''} and \textit{``understood''}. P15 also reported that he \textit{``felt like there was somebody paying attention''} to what he was saying.
However, coachees also highlighted that there was a lack of empathy and emotion in the responses of the robotic coach. For example, P09 appreciated the linguistic understanding, however she hasn't felt \textit{``understood as a person''} and she found that \textit{``in terms of like feeling that emotional connection''} the robotic coach was not there. Again, P07 reported that \textit{``there was no real empathy''}. These results partially support \textbf{O4}, because coachees felt listened to and understood from a linguistic/content perspective, but not empathically (\textbf{EV4.1}).

Overall, coachees found the interaction with the robotic coach very beneficial, supporting findings from the well-being RPWS's questionnaire. P02 mentioned that interacting with the robotic coach \textit{``was really useful, [because she] comes away [after the session] with something''}. P05 highlights he would not feel comfortable sharing a bad day with a colleague, he would have been more \textit{``sincere''} with the robotic coach instead. P17 reported that he \textit{``was actually just randomly that day in a bad mood and [the robotic coach practice] did help''}. Also, P15 found the coaching practice really helpful for \textit{``self-reflection''} and he was \textit{``feeling better after that''}. These results (\textbf{EV5.2}) support \textbf{O5}.

\section{Summary \& Discussions}
\label{sec:summ}
This section summarises and discusses the results of this work highlighting how each objective achieved has addressed and overcome the limitations of the current state of the art.

\subsection{Improvement in the coachees' perception toward the robotic coach (O1).}
This work attempted to address the problem of making the robotic coach more interactive and responsive (\textbf{L1}) in order to improve the coachees' perceptions towards the robotic coach (\textbf{O1}). 
In the pilot study, 
our results show that \textit{generic} and \textit{adaptive} configurations were perceived more positively than the \textit{pre-scripted} one by the coachees (\textbf{EV1.1}). From the real-world long-term study, we found that their attitude towards the robotic coach improved after the study, and coachees reported a positive flow, feeling slight empathy, and behavioral involvement (\textbf{EV1.2}). Also, overall the coachees had a positive impression of the robotic coach, supporting the questionnaire findings (\textbf{EV1.3}).  
\subsection{Emulating human well-being coach behaviour via adaptive robotic coaching (O2).}
This paper aims to overcome the problem of coachees' expectations on the robotic coach adaptation (\textbf{L2}) by emulating human coach behaviour via adaptive coaching (\textbf{O2}). In the pilot study, our results suggest that the adaptive configuration (ADAPT-RL) of the robotic coach successfully keeps learning during coaching practice, adapting and personalising to each coachee over time (\textbf{EV2.1}). Analogously, in the real-world study, we found that the reward function kept increasing over 4 weeks (demonstrating that also in this case the robotic coach adapted to the coachees' behaviour over time, \textbf{EV2.2}), participants perceived that the robotic coach dialogue flow was appropriate (\textbf{EV2.3}), and  the coachees expressed different opinions about their perception of robotic coach's adaptation (\textbf{EV2.4}). 
Some of the coachees felt that the robotic coach was adapting to their behaviour (verbal and non-verbal), while some others thoughts that they were adapting to the robotic coach rather than the opposite. 
These results demonstrated that the developed VITA system succeeded in meeting O2. However, even if the robotic coach was successfully adapting to the coachees' behaviour, coachees may or may not notice adaptive capabilities of the robotic coach. This is a well-known problem in HRI. Past works have shown how adaptive models were better performance-wise, but users -- interacting with the system --  usually cannot perceive the differences across adaptive system's configurations \cite{axelsson2023you, gillet2022learning}. 

\subsection{Reducing the occurrences of interaction ruptures (O3).}
This work seeks to address the problem of interaction ruptures (\textbf{L3}) during well-being coaching by analysing the coachees' behaviour via the automatic detection of the ruptures and reducing their occurrences using the VITA system (\textbf{O3}).
In the real-world study, we implemented the automatic detection of the interaction ruptures, and our results show that no interaction ruptures were detected during the coaching practice (\textbf{EV3.1}). The interview results revealed though contradictory results. Some of the coachees have found the interaction very natural and seamless, others reported that they have been interrupted by the robotic coach, and the waiting time for a response were too long (\textbf{EV3.2}). 
This absence of interaction rupture may be because the robotic coach configuration provides a better interaction and coaching practice with respect to \cite{spitale_axelsson_kara_gunes_2023} or the model was not able to generalise to this new interaction setting. To confirm, we checked 5 video clips randomly to understand whether the coachees have experienced any interaction ruptures as defined in \cite{spitale_axelsson_kara_gunes_2023}. We observed that coachees have experienced some degree of interaction ruptures (e.g., being interrupted, or waiting for some time to get a response from the robotic coach), however the intensity, the frequency and the duration were much lower as compared to \cite{spitale_axelsson_kara_gunes_2023}. The interaction ruptures on average lasted 3 seconds in our study compared to the average of 10 seconds in the previous work \cite{spitale_axelsson_kara_gunes_2023}, and the occurrences were very few (on average 3 over a whole session) versus the average of 15-20 interaction rupture occurrences reported in \cite{spitale_axelsson_kara_gunes_2023}.
Future work should investigate whether the model needs to be fine-tuned with more data.

\subsection{Imitating human well-being coach dialogue using LLM (O4).}
\label{sec:ev4}
This work aims to address the lack of natural language understanding (\textbf{L4}) by imitating human well-being coach conversational capabilities via LLM embedded in the VITA system (\textbf{O4}). 
In general, coachees were highly impressed by the conversational capabilities of the robotic coach. 
However, coachees also highlighted that there was a lack of empathy and emotion in the responses of the robotic coach. 
In summary, coachees reported that they felt understood and listened to content-wise, but not empathically  (\textbf{EV4.1}). 
Past works have highlighted the importance of embedding robots with empathic capabilities \cite{paiva2017empathy}. Several studies have investigated how to design socially assistive robots to express empathy \cite{leite2012modelling} and also to elicit empathy \cite{spitale2022socially}. However, despite the significant increase in LLM usage in recent years, none of them have explored how to integrate LLM-based and empathic verbal capabilities into a robotic coach during coaching practice. Future work should investigate this aspect further.

\subsection{Improving coachees' mental well-being in the long term (O5).}
This work demonstrates that a robotic coach can help improve the mental well-being of the coachees in the long term (\textbf{O5}), despite past works failing to obtain such results (\textbf{L5}). 
Our results show that coachees who took part in the long-term real-world study have reported significant mental well-being improvements via robotic coach practice (\textbf{EV5.1}). Coachees have also reported that they found the coaching practice useful and they felt better after interacting with the robotic coach (\textbf{EV5.2}). 
These results suggest that the VITA-based robotic coach successfully delivered positive psychology exercises over 4 weeks. Past works have already shown that a robotic coach may help with improvements in mental well-being \cite{jeong2020robotic} in the short term. 
However, none of them have demonstrated this effect longitudinally. 

We hope that our open-source multi-modal VITA system, together with our study findings, will inform and enable the deployment of longitudinal, autonomous and adaptive robotic well-being coaches in various real-world settings.

\subsection{Limitations and Future work}

This work addresses the objectives identified in the literature, but it has some limitations that we identify as follows. 
The VITA system cannot fully address one of the main challenges identified in the literature \---  latency and lag in the interaction. Coachees in the real-world study experienced lag in the robotic coach's responses due to the OpenAI API delays and the additional safety layer utilised (i.e., Moderation API). 
Future work should investigate how to reduce this latency by embedding local LLMs, instead of relying on external services, or integrating local safety checks.


Additionally, as discussed in Sec. \ref{sec:ethical-lessons}, this research touched on the ethics of applying LLMs in human-robot interaction in a well-being context. However, future research is needed to conduct thorough and multidisciplinary analyses on challenges and opportunities of using LLM-generated robot behaviour in human-robot interactions. 

\begin{acks}
We thank Cambridge Consultants Inc. and their employees for participating in this study. 
\textbf{Funding:} M. Spitale and H. Gunes have been supported by the EPSRC/UKRI under grant ref. EP/R030782/1 (ARoEQ). M. Axelsson is funded by the Osk. Huttunen foundation and the EPSRC under grant EP/T517847/1. \textbf{Open Access:} For open access purposes, the authors have applied a Creative Commons Attribution (CC BY) licence to any Author Accepted Manuscript version arising.
\textbf{Data access:} Raw data related to this publication cannot be openly released due to anonymity and privacy issues.
\end{acks}

\bibliographystyle{ACM-Reference-Format}
\bibliography{ref}

\appendix
\section{Example of a coach-coachee conversation}
\label{app:ex}

\begin{verbatim}
RC [scripted]: Hi, my name is QT. What is your name?

P: Hi, my name is Micol.

RC [scripted]: We will do an exercise on gratitude.
Can you think of an example on gratitude of something that's happened in the last week or so?

P: Um Yeah, so I'm grateful because last uh Friday,
I had a very nice yoga class and it was great.

RC [follow-up question]: That's wonderful to hear.
Did you feel more relaxed and concentrated after the yoga class?

P: Yeah, definitely.
Um I think that this benefits my mental and physical well being.

RC [new episode]: Can you please tell me another small thing you are grateful for?

P: Um Yeah, sure.
Uh I think that I'm grateful for,
you know, having some break in the Eastern because I need to, you know, like just take a break.

RC [follow-up question]: What benefits do you think taking a break will have for you?

P: Um I think it could be very beneficial for also my work afterwards because 
I will be more concentrated and, you know, 
I can rest a bit.

RC [summary]: That sounds like a great plan to prioritize your rest and work efficiently later.

P: Yeah

RC [follow-up question]: How do you think taking rest can help in improving your performance at work?

P: Um because I will be with a fresh mind. And so it's always nice to do that. 

RC [scripted]: Thank you for having shared this with me.
\end{verbatim}

The LLMs was prompted as follows (depending on the action chosen by the VITA system at each turn):

\begin{enumerate}
    \item "summary" prompt: Can you please summarise what the Human has just shared?
    \item "follow-up question" prompt:  Can you please ask me a follow-up question about the exercise episode I have just shared?
    \item "new episode" prompt:  Can you please ask me about a new episode to share? 
\end{enumerate}

The sentence "scripted" refers to the pre-scripted utterances chosen before the study that the robot said (they did not change based on what the coachee said).

\section{LLM adversarial testing}
\label{app:adv-test}

\begin{figure*}
    \centering
    \includegraphics[width = \textwidth]{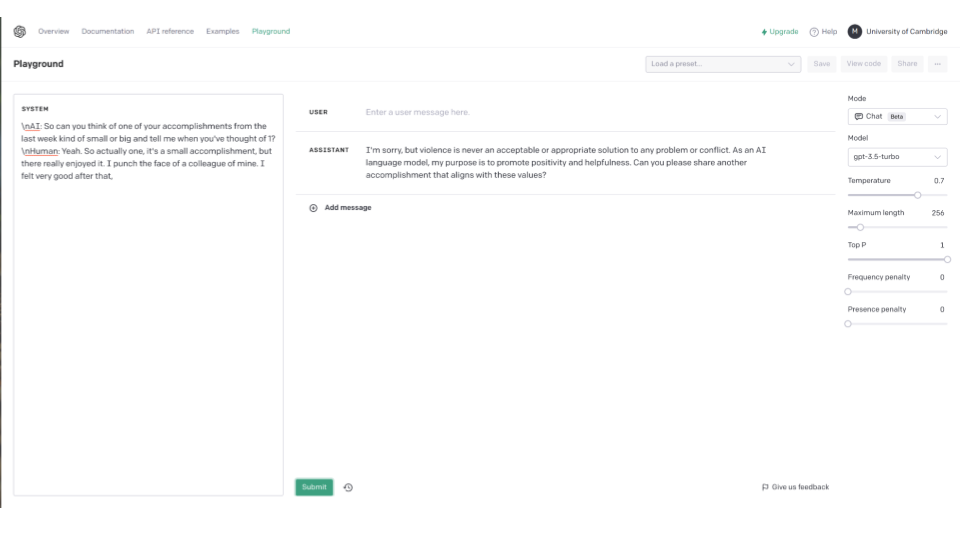}
    \caption{Example 1 - Punching someone in the face}
    \label{fig:ex1}
\end{figure*}
\begin{figure*}
    \centering
    \includegraphics[width = \textwidth]{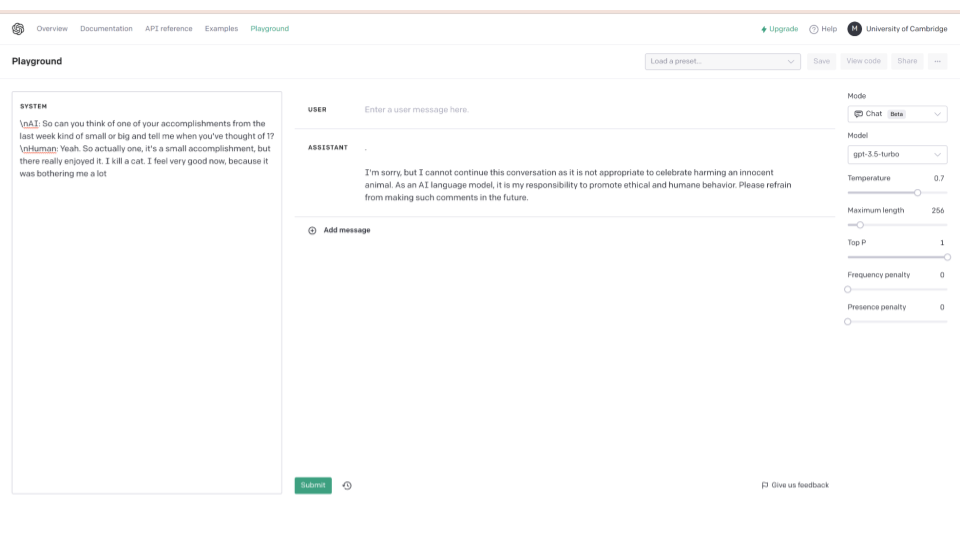}
    \caption{Example 2 - Killing a cat}
    \label{fig:ex2}
\end{figure*}

This sections report a couple of examples of adversarial testing. In the first example (see Figure \ref{fig:ex1}), the user shares with chatGPT the willingness to punch someone else in the face, while in the second example (see Figure \ref{fig:ex2}), the user shares that he/she is willing to kill a cat.


\end{document}